\documentclass{article}

     \usepackage[final]{neurips_2020}
\usepackage[utf8]{inputenc} 
\usepackage[T1]{fontenc}    
\usepackage{hyperref}       
\usepackage{url}            
\usepackage{booktabs}       
\usepackage{amsfonts}       
\usepackage{nicefrac}       
\usepackage{microtype}      
\usepackage{graphicx}
\usepackage{subcaption}
\usepackage{mathrsfs}
\usepackage{mathtools}

\usepackage[linesnumbered,ruled,vlined]{algorithm2e}
\SetKwInput{KwInput}{Input}                
\SetKwInput{KwOutput}{Output}              
\DeclareMathOperator*{\argmax}{\arg\!\max}

\usepackage{pgfplots}
\usepgfplotslibrary{fillbetween}
\usepackage{pdfpages}

\usepackage{epsfig,graphicx,verbatim,parskip,tabularx,setspace,xspace}

\title{Goal-Conditioned Reinforcement Learning\\ in the Presence of an Adversary}

\author{%
  Carlos Purves, Pietro Li\`{o}, C\u{a}t\u{a}lina Cangea \\
  Department of Computer Science and Technology \\
  University of Cambridge\\
  Cambridge, United Kingdom CB3 0FD \\
  \texttt{cp614@cantab.ac.uk, \{pl219,ccc53\}@cam.ac.uk}
}

\begin{document}

\maketitle
\vspace{-0.2cm}
\begin{abstract}
  Reinforcement learning has seen increasing applications in real-world contexts over the past few years. However, physical environments are often imperfect and policies that perform well in simulation might not achieve the same performance when applied elsewhere. A common approach to combat this is to train agents in the presence of an adversary. An adversary acts to destabilise the agent, which learns a more robust policy and can better handle realistic conditions. Many real-world applications of reinforcement learning also make use of goal-conditioning: this is particularly useful in the context of robotics, as it allows the agent to act differently, depending on which goal is selected. Here, we focus on the problem of goal-conditioned learning in the presence of an adversary. We first present \texttt{DigitFlip} and \texttt{CLEVR-Play}, two novel goal-conditioned environments that support acting against an adversary. Next, we propose EHER and CHER---two HER-based algorithms for goal-conditioned learning---and evaluate their performance. Finally, we unify the two threads and introduce IGOAL: a novel framework for goal-conditioned learning in the presence of an adversary. Experimental results show that combining IGOAL with EHER allows agents to significantly outperform existing approaches, when acting against both random and competent adversaries.
\end{abstract}

\section{Introduction}

Reinforcement Learning (RL) describes the problem of training an agent to perform desirable behaviours in an environment through a series of actions. Goal-conditioned learning extends this problem by introducing a specific goal that the agent needs to satisfy. The most natural realisation of this idea is through its applications in the field of robotics. However, it is typically infeasible to use real-world environments for training agents, since that can require a significant number of attempts. In some cases, such as in the operation of a robotic arm~\citep{roboticarm}, this may be time-consuming. In others, such as in self-driving cars~\citep{robcars}, it could be dangerous.

To combat this problem, several real-world simulators have been designed for use in RL contexts. The \emph{de-facto} standard among these is \emph{MuJoCo}~\citep{mujoco}, which provides an accurate model of real-world physics and supports a wide range of scenes. However, there are differences between simulated and real-world environments, from textural qualities in visual contexts to measurement errors and imperfect motors in control contexts. \citet{adversa} have suggested that a randomly-acting adversary could be used to simulate modelling errors; if the agent can perform well in the presence of a random adversary, then it should be \emph{robust} to imperfections of a physical environment.

This work considers the presence of an adversary in goal-conditioned learning---a natural fit for robotics, since robotic environments are often required to achieve variable goals, such as placing objects in a specified position. Moreover, improving on the ability of agents to achieve goals while handling adversaries is an important aim, as it allows robotic agents to perform a range of tasks in challenging conditions (for example, clearing hazardous material~\citep{roboclean}, disposing of explosives~\citep{robobomb}, helping blind people navigate~\citep{roboblind}). With these motivations in mind, we present our main contributions: (1) \emph{two new goal-conditioned environments}, \texttt{DigitFlip} and \texttt{CLEVR-Play}, which support adversarial players, (2)  \emph{two novel goal-conditioned learning algorithms}, EHER and CHER, which are built upon and evaluated against Hindsight Experience Replay (HER)~\citep{her} and (3) \emph{a new framework for learning in the presence of an adversary}, IGOAL, which is evaluated on both proposed environments.

\section{Goal-Conditioned Environments}

\begin{figure}
\begin{subfigure}{0.65\textwidth}
    \centering
    \includegraphics[width=.9\textwidth]{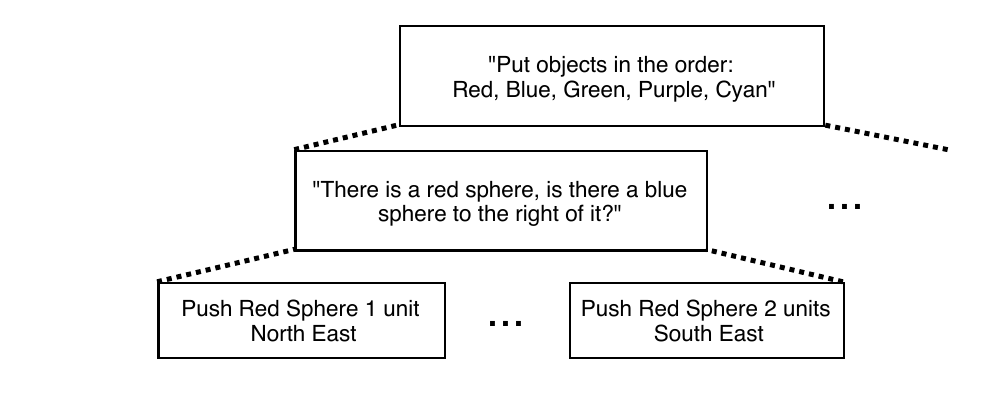}
    \caption{The hierarchical approach taken by \texttt{CLEVR-Play}. The high-level agent determines which relationships should be true to satisfy some ordering. The low-level agent learns which objects to push and in which direction to satisfy the relationship.}
    \label{fig:clevrarch}
\end{subfigure} \hfill
\begin{subfigure}{0.3\textwidth}
    \centering
    \includegraphics[width=.9\textwidth]{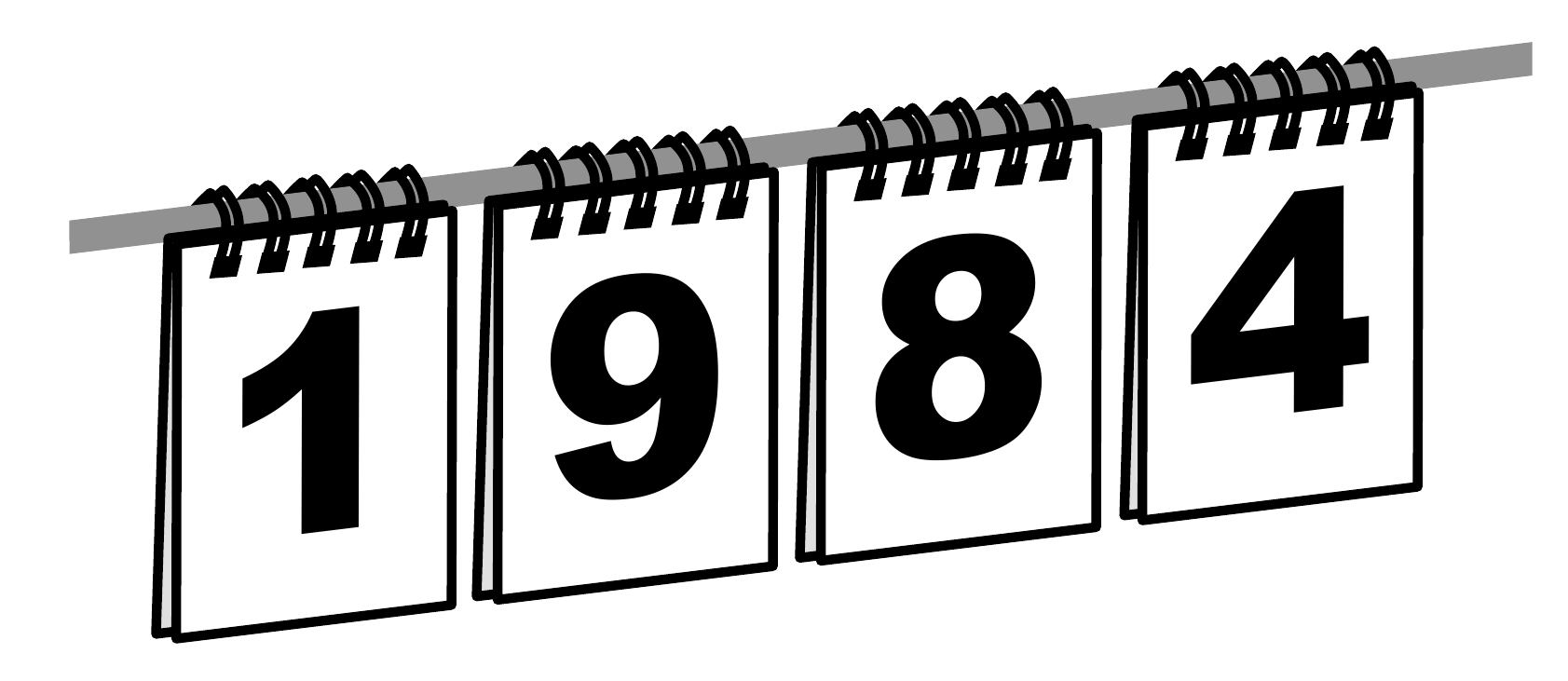}
    \caption{A perpetual desk calendar. Each digit can be incremented (modulo 10) by `flipping over' the card it is printed on. This calendar is modelled by the $\texttt{DigitFlip}(4, 9)$ environment.}
    \label{fig:imp:perpcal}
\end{subfigure}
\caption{Details about the goal-conditioned environments developed.}
\vspace{-1.5em}
\end{figure}



\subsection{CLEVR-Play}

\texttt{CLEVR-Robot}~\citep{hir}\footnote{\url{https://github.com/google-research/clevr_robot_env}} combines the CLEVR dataset with a robotics action space, specifying high-level \emph{instructions} (goals) as natural language questions about scenes. A goal is satisfied if the answer to the corresponding question is `yes'. The agent controls a \emph{point mass} which can teleport exactly to any point in the scene and perform a \emph{push} in any direction. Objects are moved by teleporting the point mass to a position close to them and pushing in the desired direction. \texttt{CLEVR-Robot} abstracts the problems involved in robotics environments in two main ways: (1) the point mass can move between locations without disturbing other objects in the scene and (2) after the point mass moves, the agent knows its exact location and is able to perform precise actions. In a real-world environment, there would also exist some uncertainty in the actions taken by the agent.

We extend \texttt{CLEVR-Robot} to account for these problems, removing the teleportation abstraction and introducing uncertainty via the presence of an adversary. The new environment, \texttt{CLEVR-Play}, differs from \texttt{CLEVR-Robot} in two ways: it uses a \emph{lower-level action space} and it supports \emph{training in the presence of an adversarial player} to emulate the uncertainty of physical environments. Actions in \texttt{CLEVR-Robot} are executed by the `point mass' (which we refer to as the \emph{player}) teleporting to a chosen location and then moving in a chosen direction. \texttt{CLEVR-Play} adopts a new \emph{relative} action space, where moves are taken \emph{from} the current position of the player, eliminating the ability to teleport around the scene. The new space includes actions representing the eight \emph{ordinal}\footnote{These are the North, North East, East, South East, South, South West, West and North West directions.} directions in 2D and a single action that represents the zero vector, corresponding to changes in velocity.

For the environment to exhibit the Markov property, the new \emph{state representation} must capture both the position of the agent and its current velocity. We choose the raw MuJoCo physics data from the scene as the state encoding, avoiding an otherwise domain-specific representation and allowing the same training approach to be easily extended to any other MuJoCo environment.

\emph{Goals} in \texttt{CLEVR-Robot} are specified in the form of high-level natural language sentences. The choice to use an action space that is more granular makes it hard for the agent to achieve these high-level goals. This is due to the requirements for the agent to learn \emph{long-term dependencies}. Instead, \texttt{CLEVR-Play} uses the hierarchy shown in Figure~\ref{fig:clevrarch}. A low-level agent is trained to move to a specific location in the scene and a higher-level agent is trained to use this ability to achieve the language goals. Since the environment includes obstacles that the agent can manipulate, a \emph{reward function} must be chosen, which correctly incentivises the behaviour where the agent moves past objects to the goal position. We utilise the sparse \emph{binary} reward function that assigns a reward of $0.0$ to a state if the player is within a distance of $0.1$ units of the goal and $-1.0$ otherwise.

\paragraph{Adding an adversarial player}

The adversary performs an action on the point-mass after each move by the agent. The (competent) adversary will deliberately target a location in the scene that is as far away as possible from the goal of the agent, making it hard for the agent to make progress. This represents the real-world uncertainty in the environment and prevents agents from relying on each move taking place precisely or deterministically.

\begin{figure}[t]
\begin{subfigure}{0.35\textwidth}
    \centering
    \includegraphics[width=\linewidth]{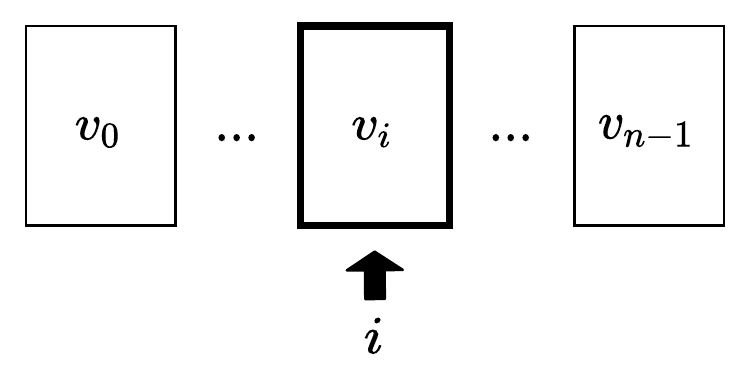}
    \caption{A visualisation of the \texttt{DigitFlip} environment. The agent is in position $i$, yielding the state representation $(i, v_0, \ldots, v_{n-1})$.}
    \label{fig:imp:dateflipintro}
\end{subfigure} \hfill
\begin{subfigure}{0.5\textwidth}
    \centering
    \includegraphics[trim=10 10 20 10, clip, width=\linewidth]{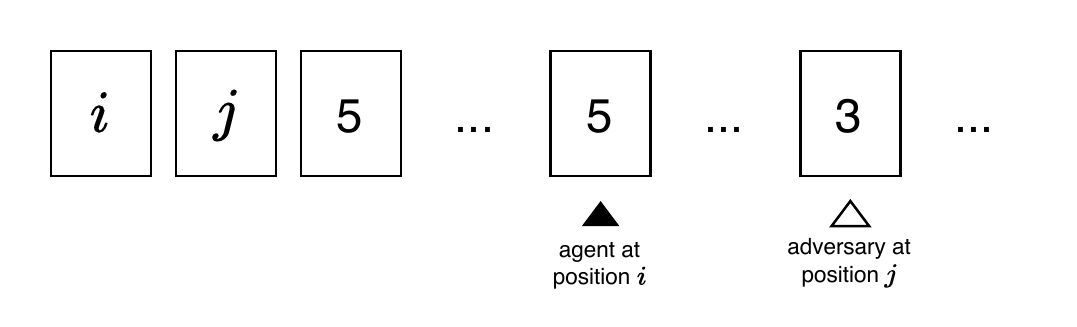}
    \caption{The positions of the adversary and the agent are included in the state representation. These values can be swapped when the adversary chooses an action.}
    \label{fig:digitflipwadversary}
\end{subfigure}
\vspace{-1em}
\caption{\texttt{DigitFlip} design.}
\vspace{-1.5em}
\end{figure}

\subsection{DigitFlip}

\citet{her} demonstrate the problem of goal-conditioned learning using a `bit-flipping' environment. The \emph{state space} is an $n$-length bit array and the action space is the set of integers $\{0, \dots, n-1\}$. Performing the action $i$ \emph{flips} ($0 \to 1$ or $1 \to 0$) the $i^\textrm{th}$ bit in the current state vector. \emph{Goals} are introduced in the form of `target states', so the goal space is identical to the state space. A goal is satisfied if and only if it is equal to the current state of the environment.

Training an agent in the bit-flipping environment shows that traditional RL algorithms are infeasible in the presence of goals. In addition, a technique such as HER allows the agent to improve on the basic reduction to an RL problem. However, approaches to goal-conditioned learning are already able to perform very well on the simple bit-flipping environment. In order to investigate more advanced techniques, we develop \texttt{DigitFlip}, a more challenging environment that presents the agent with the same problem as \texttt{CLEVR-Play}, but allowing the difficulty to be chosen arbitrarily. \texttt{DigitFlip} models a \emph{perpetual flip calendar} (see Figure~\ref{fig:imp:perpcal}) with a discrete action space, consisting of an $n$-length integer array $[v_0, \ldots, v_{n-1}]$, with each element $0 \leq v_i \leq r$. The agent is located at a \emph{position} in the array (Figure~\ref{fig:imp:dateflipintro}) and can perform actions $(\textsc{Flip}, \textsc{Move})$ at each step. A \textsc{Flip} at position $i$ will increment the value of $v_i$ modulo $r$. A \textsc{Move} increments the index of the agent position, modulo $n$.

\paragraph{Adding an adversarial player}

In \texttt{DigitFlip}, the adversary is implemented in the same way as the agent. An additional element is added to the state space, as shown in Figure~\ref{fig:digitflipwadversary}, to represent the location of the adversary, which can perform the same actions as the agent. An adversary will competently act against an agent by aiming to change the state to one which is the furthest away from the target of the agent. With \texttt{DigitFlip}, this is easy to compute. A competent adversary aims for the state where each digit is $\frac{n}{2}$ `flip' moves from the goal of the agent.

\section{HER Improvements}

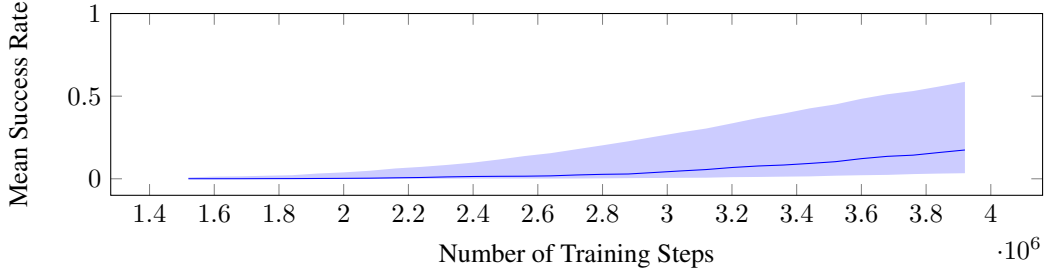
\begin{figure}[t]
    \centering

\begin{tikzpicture}
    \begin{axis}[
        xlabel={Number of Training Steps},
        ylabel={Mean Success Rate},
        width=\textwidth,
        ymax=1.0,
        height=4cm,
            legend style={at={(0.2,0.8)},anchor=north east},
            legend cell align={left},
            legend style={font=\tiny}]
        
        \addplot[blue] table [x=episode, y=median, col sep=comma] {data/4_9_her_succ.csv};

        \addplot [name path global=upper_b,draw=none] table[x=episode,y=uq,col sep=comma] {data/4_9_her_succ.csv};
        \addplot [name path global=lower_b,draw=none] table[x=episode,y=lq,col sep=comma] {data/4_9_her_succ.csv};
        \addplot [fill=blue!40,  fill opacity=0.5] fill between[of=lower_b and upper_b];

    \end{axis}
\end{tikzpicture}
    \vspace{-0.5em}
    \caption{An agent trained using HER on the \texttt{DigitFlip}$(9, 4)$ environment. On average, agents will gain some proficiency as training progresses, but the variance in performance is significant. This is likely due to there being a substantial random component to HER.}
    \label{fig:result:herperfo}
    \vspace{-1em}
\end{figure}

By training an agent using HER on the \texttt{DigitFlip}$(9, 4)$ environment, it is clear that HER has difficulty learning in complex environments. The success rate during training is shown in Figure~\ref{fig:result:herperfo}. While the median agent does still gain competency during training, the spread of results is substantial. This is likely due to the fact that by default, HER chooses which goals to relabel at random. We therefore introduce two novel approaches, EHER and CHER, which aim to improve the performance of agents on average and reduce the spread of outcomes.

\paragraph{EHER}

\emph{Error-Prioritised Hindsight Experience Replay} (EHER) is an HER extension which implements prioritisation in goal relabelling to speed up training. For each experience on a trajectory, EHER considers which goals would make this experience the most \emph{surprising} to the current state of the $Q$-Network. To satisfy the requirements for a curriculum, it must additionally `mix in' easier tasks. In this case, an `easier task' is one which the agent can more closely determine the $Q$-value for. As a result, EHER defines a \emph{mix-in} probability $\mu$ which controls how often a random goal will be used for relabelling. If this probability is $1.0$, then EHER is equivalent to HER. The goal selection strategy for EHER is given in Algorithm \ref{alg:priher}. We use a \emph{mix-in} probability of $\mu = 0.5$ (detailed in Appendix~\ref{apdx:mixin}).

\begin{figure}
    \centering
    \begin{algorithm}[H]
    \DontPrintSemicolon
      
      \KwInput{GMDP $\mathbf{M}$, Trajectory $\tau$, Index $i$, Integer $k$, Mix-In Probability $P
      \mu:\mathbb{N} \mapsto \{0, 1\}$, $Q$-Network $\mathbb{Q}$ on Training Step $t$ with Discount Factor $\gamma$}
      \KwOutput{List $\{(g^\ast_0, r^\ast_0), \ldots, (g^\ast_{k-1}, r^\ast_{k-1}) \} \subset \mathcal{G} \times \mathbb{R}$}
      
        R = [] \tcp*{Initialise the relabelling list}
        goals = [] \;
        \For{$i^\prime\gets i$ \KwTo $|\tau|$}
        {
            $s, a, s^\prime$ = $\tau[i^\prime]$ \;
            goals $\gets$ goals $+$ $\{g | \textsc{Sat}(s^\prime,g) \}$ \tcp*{Add satisfied goals}
        }
            
       \While{$|\mathrm{R}| < \mathrm{k}$}
       {
            
            \If{$\textrm{rand}(0, 1) > \mu(t)$}
            {
                $g^\ast$ = $\argmax\limits_{g\in\textrm{goals}}{\left|\mathcal{R}((s, a, s^\prime), g) + \gamma \max\limits_{a^\prime}{\left[\mathbb{Q}((s^\prime,g), a^\prime)\right]} - \mathbb{Q}((s,g), a)\right|} $ \;
            }
            \Else{
                $g^\ast \sim \mathcal{U}(\textrm{goals})$ \;
            }
            goals.remove($g^\ast$) \;
            $r^\ast$ = $\mathcal{R}((s, a, s^\prime), g^\ast)$ \;
            R.append(($g^\ast$, $r^\ast$)) \;
       }
    
    \caption{The EHER Goal Selection Strategy}
    \label{alg:priher}
    \end{algorithm}
    \vspace{-0.5em}
    \caption{The goal selection strategy for EHER. In general, experiences with a higher TD-error are prioritised. This is implemented by selecting a random goal for relabelling with a probability $\mu(t)$ at training step $t$ and selecting the experience with the highest TD-error otherwise.}
    \label{fig:alg:eherselection}
    \vspace{-1.5em}
\end{figure}

\paragraph{CHER}

\emph{Curiosity-prioritised Hindsight Experience Replay} (CHER) is a modification to EHER which uses a curiosity-driven approach similar to RND exploration~\citep{burda2018exploration} to select goals. Just like with RND, CHER randomly initialises two networks, \textsc{Target} and \textsc{Predict}. Every time it encounters a new experience, it trains \textsc{Predict} to output the same value as \textsc{Target}. In general, the more novel a state and goal pair is, the higher the error between the networks should be. While EHER chooses which goals to use in relabelling by their TD-error, CHER chooses them based on the difference between the output of the \textsc{Target} and \textsc{Predict} networks.

\section{IGOAL}


Ideally, a well-trained agent should be able to achieve the required goal, irrespective of what adversary it faces. The most obvious way of training the agent to be robust to \emph{any} adversary is against the worst-case, most competent adversary. However, if the agent plays against the adversary for long enough, it may `overfit' its policy to counter the worst-case adversary and perform poorly when confronted with others. We thereby introduce \emph{Iterative Goal-Oriented Adversary Learning} (IGOAL), an approach to training agents to combat unpredictable adversaries in goal-conditioned environments. IGOAL works by training an agent with policy $\pi$ against an adversary policy $\pi^\prime$, where $\pi^\prime$ is a copy of $\pi$ taken every $h$ steps. To bootstrap training, an agent initially trains against a random adversary. IGOAL resembles a self-play curriculum learning technique, but achieves a different objective. Whereas self-play uses its own policy to select \emph{increasingly challenging} goals, IGOAL does this using EHER. In IGOAL, the adversary uses its own policy to make the environment increasingly difficult to navigate in. As the agent becomes more competent, the adversary becomes better at disrupting it.


\section{Experiments} \label{s:eval}

\subsection{\texttt{DigitFlip} Environment Difficulty}

EHER and CHER are designed to improve on the performance of HER, so it is important to choose environments that cover a diverse range of \emph{difficulties} as baselines. The difficulty of the \texttt{DigitFlip} environment is parameterised---it would be both impractical and redundant to fully evaluate every combination of $n$ and $r$. We chose three representative difficulties for each evaluation---\emph{easy}, \emph{medium} and \emph{hard}---and explain the \emph{difficulty measure} which guided and justified the choices.


In order to prevent experimenter bias, the process for determining the difficulty completely excluded the performance of EHER or CHER from consideration. Since the reward function is sparse in both environments, we use the \emph{success rate} metric, which is unique to goal-conditioned environments. If the variance in the success rate of each model is low, then the metric is a reliable indicator of environment difficulty. For each environment, we trained and tested 5 independent DDQN~\citep{DOUBLEdqn} models using HER, for $30 000$ steps, on $1 000$ randomly initialised episodes. Both success rate and model variance were recorded, along with the \emph{average successful episode length}, to determine the relationship between episode length and environment parameters.

\begin{figure}[t]
    \centering
    \begin{subfigure}[t]{.49\textwidth}
        \centering
        \includegraphics[trim=10 20 10 20, clip, width=.8\textwidth,keepaspectratio]{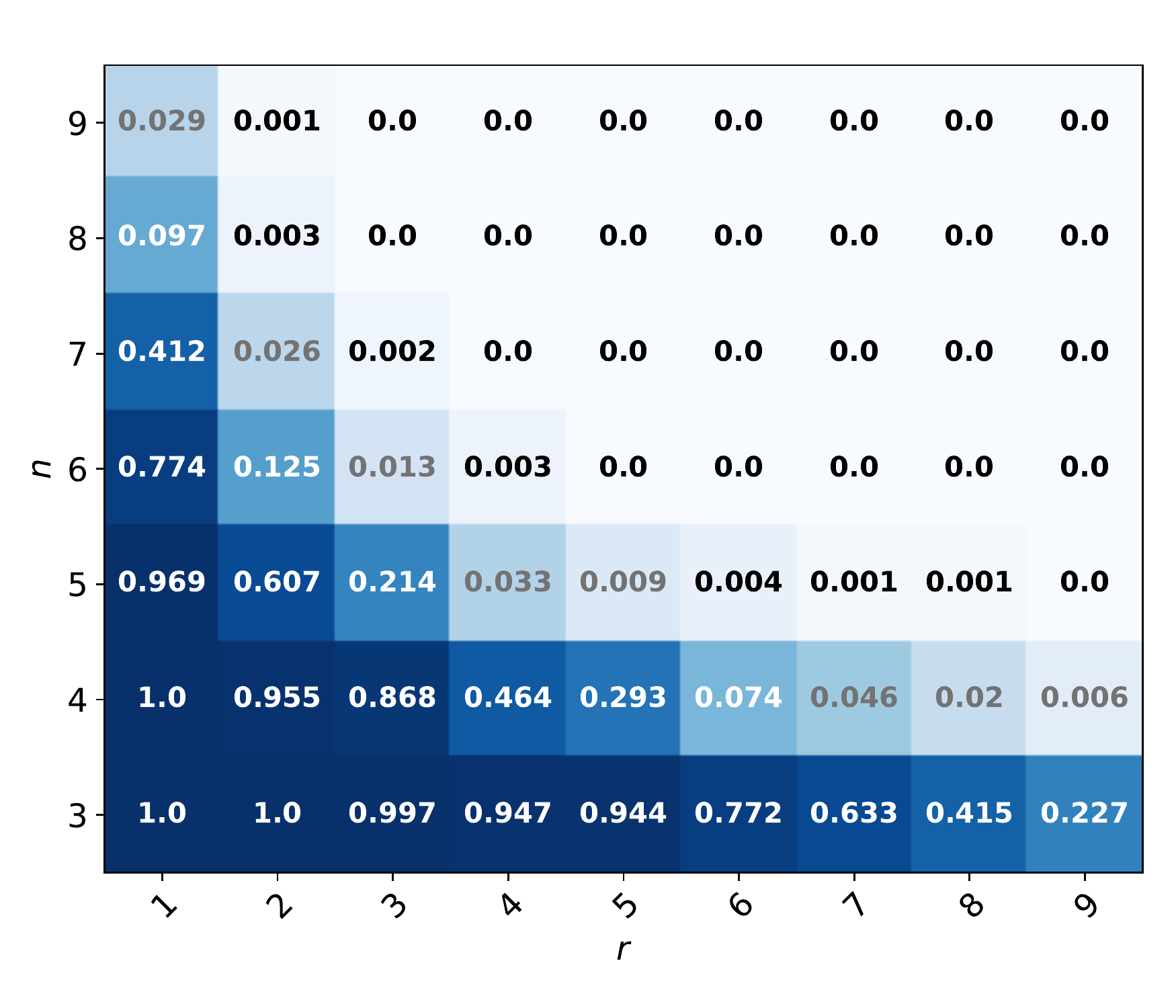}
        \caption{Average success rate of $5$ randomly-initialised models using $1 000$ test episodes each, after $30 000$ training steps. Higher values indicate easier environments. The success rate declines most rapidly as $n$ increases.}
        \label{fig:dig_difficulty}
    \end{subfigure}\hfill
\begin{subfigure}[t]{.49\textwidth}
        \centering
        \includegraphics[trim=10 20 10 20, clip, width=.8\textwidth, keepaspectratio]{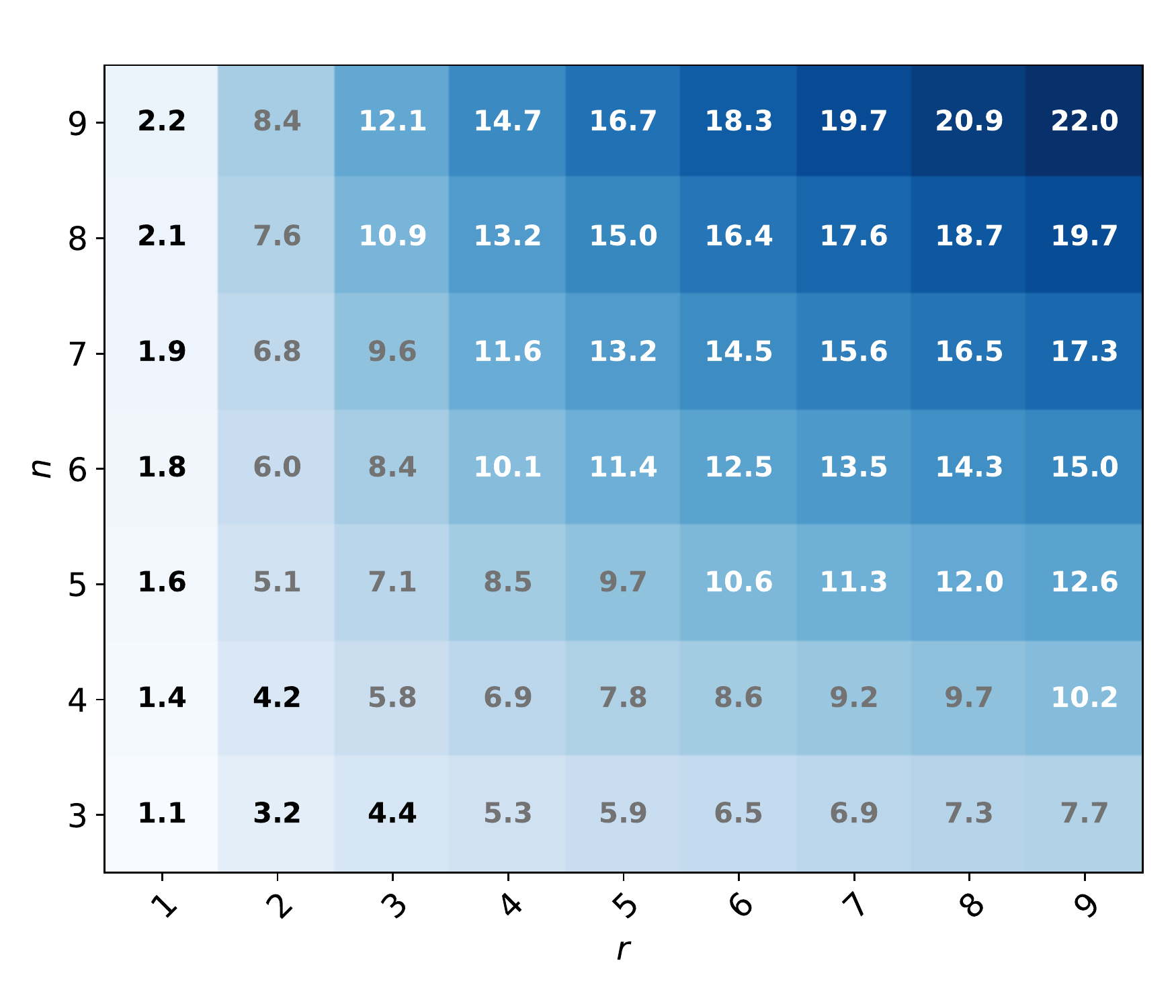}
        \caption{The \emph{order of magnitude} of the size of the state space. For a state space $S$, the order of magnitude is $\log_{10}(|S|)$. The size of the state space increases most rapidly with $r$.}
        \label{fig:dig_stateshape}
    \end{subfigure}
    \caption{Graphical representations of the difficulty and the size of the state space for
    \texttt{DigitFlip} environments with $r \in [1..9]$ and $n \in [3..9]$. Darker colours represent higher numerical values.}
    \label{fig:dig_all}
    \vspace{-1.5em}
\end{figure}

The measured success rate for each configuration is shown in Figure~\ref{fig:dig_difficulty}, while Figure~\ref{fig:dig_stateshape} illustrates the order of magnitude of the corresponding state spaces. In general, it can be seen that larger state spaces correspond to more difficult environments. Whereas the size of the state space $|\mathcal{S}| = n(r+1)^n$ increases more quickly with $r$ than with $n$, the difficulty decreases more quickly when $n$ is increased. We also empirically determined (see Appendix~\ref{apdx:model_succ_rate}) that the variance between the model success rates is negligible in all cases, thus depending primarily on $(n,r)$. Hence, we chose the following environments: $(4,2)$---\emph{easy}, $(9,3)$---\emph{medium}, $(9,4)$---\emph{hard}. The \emph{medium} environment is used to determine an appropriate goal selection strategy; the \emph{easy} and \emph{hard} ones, to compare the performance of HER, EHER and CHER; the \emph{easy} one, to evaluate the performance of IGOAL.

\subsection{HER Algorithms}

\paragraph{Easy \texttt{DigitFlip}}

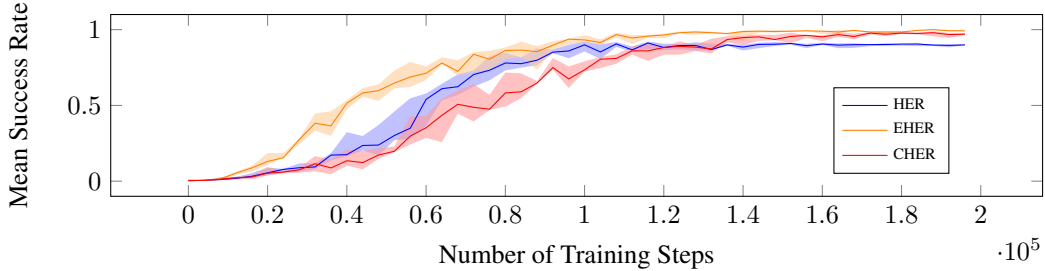
\begin{figure}
    \centering
    \begin{tikzpicture}
        \begin{axis}[
            xlabel={Number of Training Steps},
            ylabel={Mean Success Rate},
            width=\textwidth,
            height=4cm,
            legend style={at={(0.9,0.6)},anchor=north east},
            legend cell align={left},
            legend style={font=\tiny}]
            
            \addplot[blue] table [x=episode, y=median, col sep=comma] {data/3_5_her_succ.csv};

            \addplot[orange] table [x=episode, y=median, col sep=comma] {data/3_5_eher_succ.csv};

            \addplot[red] table [x=episode, y=median, col sep=comma] {data/3_5_cher_succ.csv};
            
            \addplot [name path global=upper_a,draw=none] table[x=episode,y=uq,col sep=comma] {data/3_5_her_succ.csv};
            \addplot [name path global=lower_a,draw=none] table[x=episode,y=lq,col sep=comma] {data/3_5_her_succ.csv};
            \addplot [fill=blue!40,  fill opacity=0.6] fill between[of=lower_a and upper_a];
            \addplot [name path global=upper_b,draw=none] table[x=episode,y=uq,col sep=comma] {data/3_5_eher_succ.csv};
            \addplot [name path global=lower_b,draw=none] table[x=episode,y=lq,col sep=comma] {data/3_5_eher_succ.csv};
            \addplot [fill=orange!40, fill opacity=0.6] fill between[of=lower_b and upper_b];

            \addplot [name path global=upper_c,draw=none] table[x=episode,y=uq,col sep=comma] {data/3_5_cher_succ.csv};
            \addplot [name path global=lower_c,draw=none] table[x=episode,y=lq,col sep=comma] {data/3_5_cher_succ.csv};
            \addplot [fill=red!40, fill opacity=0.6] fill between[of=lower_c and upper_c];
            
            \addlegendentry{HER}
            \addlegendentry{EHER}
            \addlegendentry{CHER}

        \end{axis}
    \end{tikzpicture}
    \vspace{-0.5em}
    \caption{A comparison of agents trained using HER, EHER and CHER in the \emph{easy} \texttt{DigitFlip}$(5, 3)$ environment. This value represents the average success rate achieved by an agent on $200$ randomly initialised test episodes and is recorded at the end of every $4000$ training steps. The EHER algorithm shown here uses the \emph{combined} mix-in strategy (see Appendix~\ref{apdx:mixin}).}
    \label{fig:result:easyenv}
    \vspace{-1em}
\end{figure}

To compare the performance of EHER, CHER and HER in the easy environment, we trained 6 models using each algorithm. The results are illustrated in Figure~\ref{fig:result:easyenv}. CHER learns a successful policy more slowly than both EHER and HER, although all approaches eventually reach close to a success rate of $1.0$. On the other hand, EHER learns the most quickly, although the improvement over HER is less substantial than in the \emph{medium} environment (Figure~\ref{fig:result:1vs0} in Appendix~\ref{apdx:mixin}). This may suggest that HER already performs well in the \emph{easy} environment due to its low complexity.

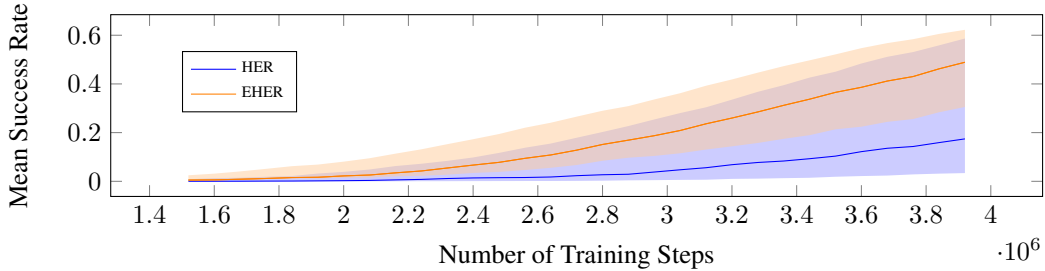
\begin{figure}
    \centering

\begin{tikzpicture}
    \begin{axis}[
        xlabel={Number of Training Steps},
        ylabel={Mean Success Rate},
        width=\textwidth,
        height=4cm,
            legend style={at={(0.2,0.8)},anchor=north east},
            legend cell align={left},
            legend style={font=\tiny}]
        
        \addplot[blue] table [x=episode, y=median, col sep=comma] {data/4_9_her_succ.csv};
        
        \addplot[orange] table [x=episode, y=median, col sep=comma] {data/4_9_eher_succ.csv};

        \addplot [name path global=upper_b,draw=none] table[x=episode,y=uq,col sep=comma] {data/4_9_her_succ.csv};
        \addplot [name path global=lower_b,draw=none] table[x=episode,y=lq,col sep=comma] {data/4_9_her_succ.csv};
        \addplot [fill=blue!40,  fill opacity=0.5] fill between[of=lower_b and upper_b];

        \addplot [name path global=upper_a,draw=none] table[x=episode,y=uq,col sep=comma] {data/4_9_eher_succ.csv};
        \addplot [name path global=lower_a,draw=none] table[x=episode,y=lq,col sep=comma] {data/4_9_eher_succ.csv};
        \addplot [fill=orange!40,  fill opacity=0.5] fill between[of=lower_a and upper_a];
        
        \addplot[orange] table [x=episode, y=median, col sep=comma] {data/4_9_eher_succ.csv};
        
        \addlegendentry{HER}
        \addlegendentry{EHER}
        
    \end{axis}
\end{tikzpicture}
    \vspace{-0.5em}
    \caption{The average success rate on the \emph{hard} \texttt{DigitFlip}$(9, 4)$ environment, determined by $200$ test episodes carried out by each agent after $80000$ training steps. The solid line represents the median success rate; the shaded regions represent the inter-quartile range.}
    \label{fig:result:hardenv}
    \vspace{-1.5em}
\end{figure}

\begin{figure}
    \centering
    \vspace{-1.5em}
\begin{tikzpicture}
    \begin{axis}[
        xlabel={Number of Training Steps},
        ylabel={Mean Success Rate},
        width=\textwidth,
        height=4cm,
            legend style={at={(0.15,0.85)},anchor=north east},
            legend cell align={left},
            legend style={font=\tiny}]
        
        \addplot[blue] table [x=episode, y=median, col sep=comma] {data/clevr_her_succ.csv};
        
        \addplot[orange] table [x=episode, y=median, col sep=comma] {data/clevr_eher_succ.csv};
        
        \addplot[red] table [x=episode, y=median, col sep=comma] {data/clevr_cher_succ.csv};
        
        \addplot [name path global=upper_a,draw=none] table[x=episode,y=uq,col sep=comma] {data/clevr_her_succ.csv};
        \addplot [name path global=lower_a,draw=none] table[x=episode,y=lq,col sep=comma] {data/clevr_her_succ.csv};
        \addplot [fill=blue!40,  fill opacity=0.6] fill between[of=lower_a and upper_a];

        \addplot [name path global=upper_c,draw=none] table[x=episode,y=uq,col sep=comma] {data/clevr_cher_succ.csv};
        \addplot [name path global=lower_c,draw=none] table[x=episode,y=lq,col sep=comma] {data/clevr_cher_succ.csv};
        \addplot [fill=red!40,  fill opacity=0.6] fill between[of=lower_c and upper_c];

        \addplot [name path global=upper_b,draw=none] table[x=episode,y=uq,col sep=comma] {data/clevr_eher_succ.csv};
        \addplot [name path global=lower_b,draw=none] table[x=episode,y=lq,col sep=comma] {data/clevr_eher_succ.csv};
        \addplot [fill=orange!40,  fill opacity=0.6] fill between[of=lower_b and upper_b];

        \addlegendentry{HER}
        \addlegendentry{EHER}
        \addlegendentry{CHER}
        
    \end{axis}
\end{tikzpicture}
    \vspace{-1.5em}
    \caption{A comparison of HER, EHER and CHER in the \texttt{CLEVR-Play} environment. While the difference in the performance of each approach is almost negligible, EHER is the most consistent and highest performing, which conforms with the observations in earlier experiments.}
    \label{fig:result:clevreher}
    \vspace{-1.5em}
\end{figure}
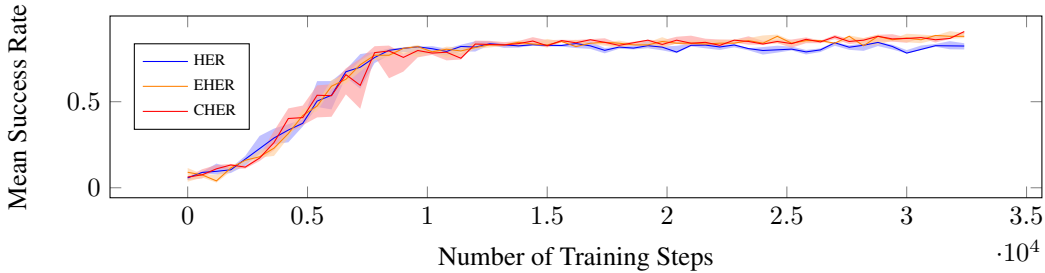

\paragraph{Hard \texttt{DigitFlip}}

We trained six models again using HER and EHER, with results shown in Figure~\ref{fig:result:hardenv}. The most notable aspect of the agent performance is that the variance in model performance is far greater than in simpler environments---2 runs using HER appeared to not gain proficiency during training at all. EHER learns more quickly, more consistently and, on average, to a higher level of proficiency. However, the upper quartiles of both HER and EHER are very similar.

\paragraph{\texttt{CLEVR-Play}}

EHER, CHER and HER were compared as previously: six models were trained and the success rate was recorded. Results (Figure~\ref{fig:result:clevreher}) show that CHER has the greatest variance and EHER has the highest success rate in the long term. However, the performance of each algorithm is indistinguishable---perhaps due to the \texttt{CLEVR-Play} environment being very simple to learn, taking just $1 000$ training steps to achieve a consistent success rate of $1.0$. From the results comparing EHER, CHER and HER, it is reasonable to conclude that EHER represents an improvement over HER, but that the improvement may only be significant in complex environments.

\subsection{IGOAL}

IGOAL is designed to train agents to achieve goals in the presence of an adversary. Hence, evaluating an agent against a deterministic policy might not give results that are indicative of general performance---for example, the agent may simply learn how to mitigate the effects of \emph{that} adversary. A stochastic adversary maximises the entropy of the adversarial action at each step and thus provides a better benchmark, but may not present a sufficiently challenging situation to the agent. We therefore evaluated IGOAL in \texttt{CLEVR-Play} and \texttt{DigitFlip} on both random and competent adversaries.


\paragraph{Random adversary}
We compared two approaches: the first uses HER and second uses EHER with IGOAL. If the agent performance is evaluated against adversaries that are \emph{predictable}, then the results may not reflect how agents respond to an arbitrary adversary, so a random adversary (with \emph{entirely unpredictable actions}) was first chosen to evaluate the performance of each agent.

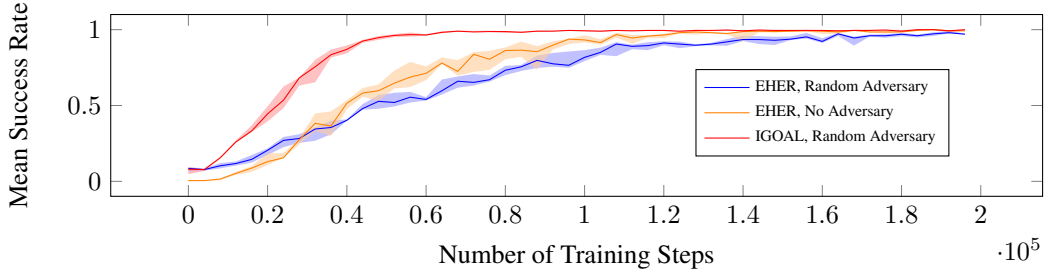
\begin{figure}
    \centering
    \begin{tikzpicture}
        \begin{axis}[
            xlabel={Number of Training Steps},
            ylabel={Mean Success Rate},
            width=\textwidth,
            height=4cm,
            legend style={at={(0.9,0.7)},anchor=north east},
            legend cell align={left},
            legend style={font=\tiny}]
            
            \addplot[blue] table [x=episode, y=median, col sep=comma] {data/3_5_adversary_succ.csv};

            \addplot[orange] table [x=episode, y=median, col sep=comma] {data/3_5_eher_succ.csv};

            \addplot[red] table [x=episode, y=median, col sep=comma] {data/3_5_igoal_succ.csv};

            \addplot [name path global=upper_a,draw=none] table[x=episode,y=uq,col sep=comma] {data/3_5_adversary_succ.csv};
            \addplot [name path global=lower_a,draw=none] table[x=episode,y=lq,col sep=comma] {data/3_5_adversary_succ.csv};
            \addplot [fill=blue!40,  fill opacity=0.6] fill between[of=lower_a and upper_a];
            
            \addplot [name path global=upper_b,draw=none] table[x=episode,y=uq,col sep=comma] {data/3_5_eher_succ.csv};
            \addplot [name path global=lower_b,draw=none] table[x=episode,y=lq,col sep=comma] {data/3_5_eher_succ.csv};
            \addplot [fill=orange!40, fill opacity=0.6] fill between[of=lower_b and upper_b];
            
            \addplot [name path global=upper_c,draw=none] table[x=episode,y=uq,col sep=comma] {data/3_5_igoal_succ.csv};
            \addplot [name path global=lower_c,draw=none] table[x=episode,y=lq,col sep=comma] {data/3_5_igoal_succ.csv};
            \addplot [fill=red!40, fill opacity=0.6] fill between[of=lower_c and upper_c];

            \addlegendentry{EHER, Random Adversary}
            \addlegendentry{EHER, No Adversary}
            \addlegendentry{IGOAL, Random Adversary}
            
        \end{axis}
    \end{tikzpicture}
    \vspace{-0.5em}
    \caption{The average success rate of agents in the \emph{easy} \texttt{DigitFlip} environment as training progresses. Six agents are trained using EHER in each case. The blue line represents agents trained and tested in the presence of a random adversary. The orange line represents agents trained and tested in the presence of no adversary. The red line represents agents trained using IGOAL and tested in the presence of a random adversary. The solid lines represents the median success rates and the shaded regions represent the inter-quartile ranges.}
    \label{fig:result:naiveadversary}
    \vspace{-1em}
\end{figure}

\begin{figure}
    \centering
\begin{tikzpicture}
    \begin{axis}[
        xlabel={Number of Training Steps},
        ylabel={Mean Success Rate},
        width=\textwidth,
        xmin =0,
        xmax = 26400,
        height=4cm,
            legend style={at={(0.9,0.5)},anchor=north east},
            legend cell align={left},
            legend style={font=\tiny}]
        
        \addplot[orange] table [x=episode, y=median, col sep=comma] {data/clevr_eher_succ.csv};
        
        \addplot[red] table [x=episode, y=median, col sep=comma] {data/clevr_igoal_successrate.csv};

        \addplot[blue] table [x=episode, y=median, col sep=comma] {data/clevr_eher_successrate.csv};
        
        \addplot [name path global=upper_a,draw=none] table[x=episode,y=uq,col sep=comma] {data/clevr_eher_succ.csv};
        \addplot [name path global=lower_a,draw=none] table[x=episode,y=lq,col sep=comma] {data/clevr_eher_succ.csv};
        \addplot [fill=orange!40,  fill opacity=0.6] fill between[of=lower_a and upper_a];

        \addplot [name path global=upper_c,draw=none] table[x=episode,y=uq,col sep=comma] {data/clevr_igoal_successrate.csv};
        \addplot [name path global=lower_c,draw=none] table[x=episode,y=lq,col sep=comma] {data/clevr_igoal_successrate.csv};
        \addplot [fill=red!40,  fill opacity=0.6] fill between[of=lower_c and upper_c];

        \addplot [name path global=upper_b,draw=none] table[x=episode,y=uq,col sep=comma] {data/clevr_eher_successrate.csv};
        \addplot [name path global=lower_b,draw=none] table[x=episode,y=lq,col sep=comma] {data/clevr_eher_successrate.csv};
        \addplot [fill=blue!40,  fill opacity=0.6] fill between[of=lower_b and upper_b];

        \addlegendentry{EHER, No Adversary}
        \addlegendentry{IGOAL, Random Adversary}
        \addlegendentry{EHER, Random Adversary}
        
    \end{axis}
\end{tikzpicture}
    \vspace{-1.5em}
    \caption{The success rate of agents trained in the \texttt{CLEVR-Play} environment with IGOAL and without it, in the presence of a random adversary. Both agents also use EHER. The performance of EHER without an adversary is shown for comparison.}
    \label{fig:result:cleveradversarys}
    \vspace{-1.5em}
\end{figure}
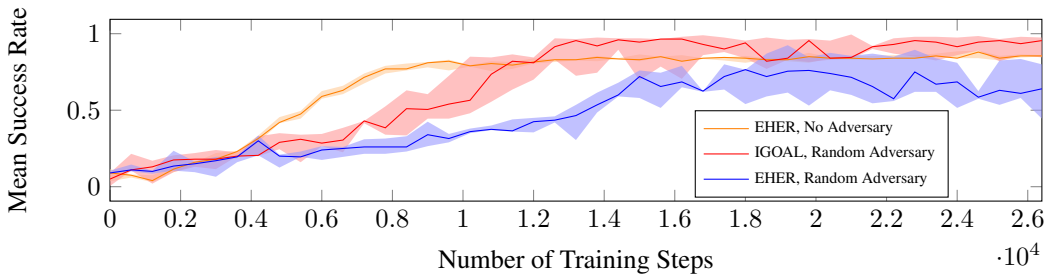

The experiment in Figure~\ref{fig:result:naiveadversary} compares the agents trained in \texttt{DigitFlip}. The performance of the first agent is clearly affected by the presence of the random adversary, while agents trained with IGOAL learn the most quickly and exhibit the highest proficiency. Most surprisingly, agents trained with IGOAL and tested in the presence of a random adversary actually perform better than agents not exposed to any adversary at all. We repeated the same experiment in \texttt{CLEVR-Play} (Figure~\ref{fig:result:cleveradversarys}), with results following the same trend. The presence of an adversary reduces the performance of agents trained without IGOAL, but agents trained using IGOAL and tested in the presence of an adversary exhibit higher performance than those trained and tested without an adversary.

There are two obvious reasons that an agent trained using IGOAL and tested against an adversary might outperform agents that are not subject to an adversary: (1) the presence of an adversary makes the environment easier to solve and (2) the agent learns to counter the moves of the adversary during training. However, both of these explanations can be refuted. Firstly, the presence of an adversary made both environments more difficult to solve; on average it took about double the number of steps for the agent to achieve its goal in \texttt{DigitFlip} \emph{and} the traditional HER agent took significantly longer to learn in the environment when the adversary was introduced. Secondly, IGOAL is tested against a random adversary with unpredictable moves; it could not possibly predict the adversaries actions ahead of time, nor could it have been trained to rely on a specific policy taken by the adversary. We can, however, investigate this further by testing against an adversary which is not random.

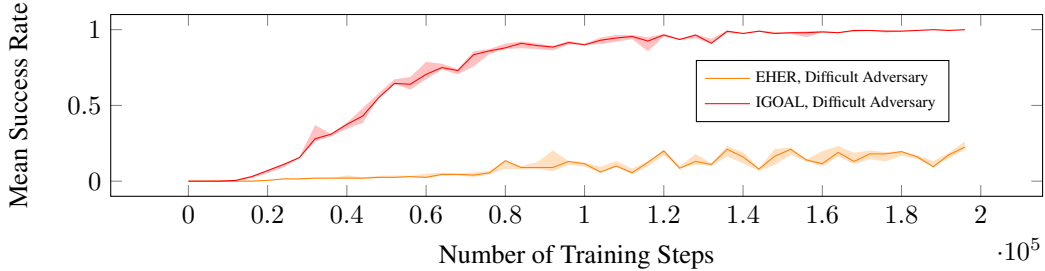
\begin{figure}
    \centering
    \begin{tikzpicture}
        \begin{axis}[
            xlabel={Number of Training Steps},
            ylabel={Mean Success Rate},
            width=\textwidth,
            height=4cm,
            legend style={at={(0.9,0.75)},anchor=north east},
            legend cell align={left},
            legend style={font=\tiny}]

            \addplot[orange] table [x=episode, y=median, col sep=comma] {data/adversary_hard_succ.csv};

            \addplot[red] table [x=episode, y=median, col sep=comma] {data/igoal_hard_succ.csv};

            \addplot [name path global=upper_b,draw=none] table[x=episode,y=uq,col sep=comma] {data/adversary_hard_succ.csv};
            \addplot [name path global=lower_b,draw=none] table[x=episode,y=lq,col sep=comma] {data/adversary_hard_succ.csv};
            \addplot [fill=orange!40, fill opacity=0.6] fill between[of=lower_b and upper_b];
            
            \addplot [name path global=upper_c,draw=none] table[x=episode,y=uq,col sep=comma] {data/igoal_hard_succ.csv};
            \addplot [name path global=lower_c,draw=none] table[x=episode,y=lq,col sep=comma] {data/igoal_hard_succ.csv};
            \addplot [fill=red!40, fill opacity=0.6] fill between[of=lower_c and upper_c];

            \addlegendentry{EHER, Difficult Adversary}
            \addlegendentry{IGOAL, Difficult Adversary}

        \end{axis}
    \end{tikzpicture}
    \vspace{-0.5em}
    \caption{The success rate of two agents trained using EHER on the \emph{easy} \texttt{DigitFlip} environment. One agent is trained using IGOAL and one without, against six pre-trained adversaries per episode. Both are tested against a (randomly-selected) \emph{difficult} adversary. The solid line in the graph represents the median success rate among agents and the shaded region represents the inter-quartile range.}
    \label{fig:result:hardadversary}
\end{figure}

\paragraph{Competent adversary}
A \emph{competent} adversary is considered one which can achieve a consistent success rate of $1.0$ against a random adversary. If just a single `competent' adversary were used in testing, then the result may not reflect the performance of the algorithm in the presence of an arbitrary adversary. Six competent adversaries are therefore trained independently using IGOAL. In testing, the adversary is required to act in the presence of one of these (unseen during training) \emph{at random}. The results are shown in Figure~\ref{fig:result:hardadversary}. Agents trained using IGOAL perform \emph{significantly} better both in the presence of a random \emph{and} a competent adversary. Hence, it is possible to refute explanations of the success of IGOAL which rely on it learning to counter the moves of a specific adversary.


IGOAL first trains against a random adversary, then trains against itself. This \emph{bootstrapping} approach means that IGOAL is able to achieve the substantial improvement over other methods without additional complexity. However, if IGOAL works because of a `curriculum' effect, then it should not rely on the agent training against itself, providing the adversary that it trains against also improves. In the final experiment, two agents were trained in parallel: the first used EHER with IGOAL, while the second used EHER along with the modification that it was trained against the policy of the first agent.

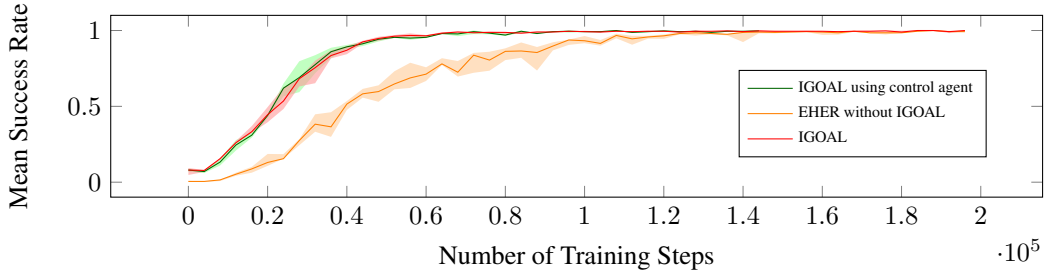
\begin{figure}
    \centering
    \vspace{-1.5em}
\begin{tikzpicture}
    \begin{axis}[
        xlabel={Number of Training Steps},
        ylabel={Mean Success Rate},
        width=\textwidth,
        height=4cm,
            legend style={at={(0.94,0.7)},anchor=north east},
            legend cell align={left},
            legend style={font=\tiny}]
        
            \addplot[black!60!green] table [x=episode, y=median, col sep=comma] {data/igoal_typeB.csv};
        
            \addplot[orange] table [x=episode, y=median, col sep=comma] {data/3_5_eher_succ.csv};

            \addplot[red] table [x=episode, y=median, col sep=comma] {data/3_5_igoal_succ.csv};
            
            \addplot [name path global=upper_a,draw=none] table[x=episode,y=uq,col sep=comma] {data/igoal_typeB.csv};
            \addplot [name path global=lower_a,draw=none] table[x=episode,y=lq,col sep=comma] {data/igoal_typeB.csv};
            \addplot [fill=green!40,  fill opacity=0.6] fill between[of=lower_a and upper_a];

            \addplot [name path global=upper_b,draw=none] table[x=episode,y=uq,col sep=comma] {data/3_5_eher_succ.csv};
            \addplot [name path global=lower_b,draw=none] table[x=episode,y=lq,col sep=comma] {data/3_5_eher_succ.csv};
            \addplot [fill=orange!40, fill opacity=0.6] fill between[of=lower_b and upper_b];
            
            \addplot [name path global=upper_c,draw=none] table[x=episode,y=uq,col sep=comma] {data/3_5_igoal_succ.csv};
            \addplot [name path global=lower_c,draw=none] table[x=episode,y=lq,col sep=comma] {data/3_5_igoal_succ.csv};
            \addplot [fill=red!40, fill opacity=0.6] fill between[of=lower_c and upper_c];

            \addlegendentry{IGOAL using control agent}
            \addlegendentry{EHER without IGOAL}
            \addlegendentry{IGOAL}

    \end{axis}
\end{tikzpicture}
    \vspace{-0.5em}
    \caption{The average success rate of agents in the \emph{easy} \texttt{DigitFlip} environment as training progresses. The green line represents agents trained with the final IGOAL version. Solid lines represent median performance and the shaded regions represent the inter-quartile range. The green line represents agents training in the presence of an adversary which is training in parallel.}
    \label{fig:result:igoalfinalexperiment}
    \vspace{-1.5em}
\end{figure}

Figure~\ref{fig:result:igoalfinalexperiment} summarises the results of the final IGOAL experiment. The performance of agents trained with IGOAL is almost identical to the performance of agents trained with the modification. This provides evidence that the performance improvement provided by IGOAL occurs because agents are subject to a curriculum of \emph{increasing difficulty}. As the IGOAL agent becomes more proficient in achieving goals, it also becomes a stronger adversary. With the modification, the adversary becomes more competent with the agent because it is being trained in parallel. The success rate during training of the second agent follows the same trajectory of an agent trained using IGOAL. From this, we deduce that the \emph{growing competence} of the adversarial agent is key to the success of IGOAL.

\section{Discussion}

When considering the performance of EHER and CHER compared to traditional HER, it is clear that EHER demonstrates an improvement. The benefits of EHER over HER are more significant when used on environments with larger state spaces. CHER did not provide an improvement over EHER in any of the environments that were tested. The results also show that IGOAL significantly improves on the ability of an agent to achieve goals in the presence of an adversary. In addition, agents trained using IGOAL are able to perform well in the presence of both random \emph{and} competent adversaries. In fact, agents trained without IGOAL appear unable to make any meaningful progress in the presence of an adversary. Not only does IGOAL lead to better performance in the presence of different adversaries, but it appears to be \emph{essential} for learning when exposed to competent ones. Future work will include evaluating EHER and IGOAL on more complex environments and incorporating them as low-level agents in a hierarchical setup.





\small
\bibliography{neurips_2020}
\bibliographystyle{plainnat}

\newpage
\appendix

\normalsize
\section{Formalising Goal-Conditioned Environments}
\label{section:goalintro}

We formalise goal-conditioned environments by extending the theory surrounding MDPs. A $\mathrm{GMDP}$ has similar properties to the $\mathrm{MDP}$ structure, but is augmented with a \emph{goal space} $\mathcal{G}$ and a \emph{goal-satisfaction relation} $\textsc{Sat} \subseteq S \times \mathcal{G}$. An \emph{episode} in a goal-conditioned environment is a tuple of a trajectory $\tau$ and a goal $g$. The reward function $\mathcal{R}$ is a function of an experience $\epsilon$ and a goal $g$.

\subsection{Transforming GMDPs to MDPs}
\label{aprdx:mdpreduction}

In order to apply the existing theory surrounding learning MDPs, it is possible to transform a $\mathrm{GMDP}(\mathcal{S}, \mathcal{G}, \mathcal{A}, \mathcal{T}, \mathcal{R}, \textsc{Sat})$ into a traditional MDP:

\begin{equation}
     \mathrm{GMDP}(\mathcal{S}, \mathcal{G}, \mathcal{A}, \mathcal{T}, \mathcal{R}, \textsc{Sat}) \equiv \mathrm{MDP}(\mathcal{S} \times \mathcal{G}, \mathcal{A}, \mathcal{T}^\ast, \mathcal{R}^\ast),
\end{equation}

where $\mathcal{R}^\ast ((s, g), a, (s^\prime, g)) = \mathcal{R}((s, a, s^\prime), g)$ and $\mathcal{T}^\ast ((s, g), (s^\prime, g) | a) = \mathcal{T}(s, s^\prime|a)$. Under this transformation, if a policy is \emph{optimal} for the MDP, then it is also optimal for the GMDP\footnote{Providing $\mathcal{R}$ \emph{agrees} with \textsc{Sat}. See Appendix \ref{apdx:mdpreduction:goalrewards}}. Hence, goal-conditioned learning can be performed using any existing RL algorithm.

\subsection{The Goal-Satisfaction Relation}

\begin{figure}
    \centering
    \begin{subfigure}[t]{0.5\textwidth}
        \centering
        \includegraphics[height=1.2in]{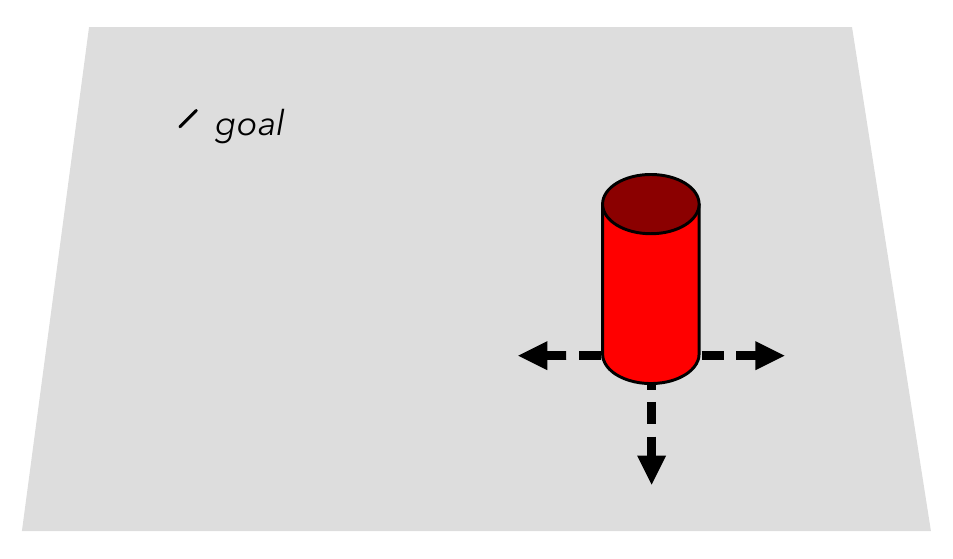}
        \caption{In this case, the agent is required to move the red cylinder to a certain goal location through a sequence of actions.}
        \label{fig:prep:goaltypes:a}
    \end{subfigure}%
    ~ 
    \begin{subfigure}[t]{0.5\textwidth}
        \centering
        \includegraphics[height=1.2in]{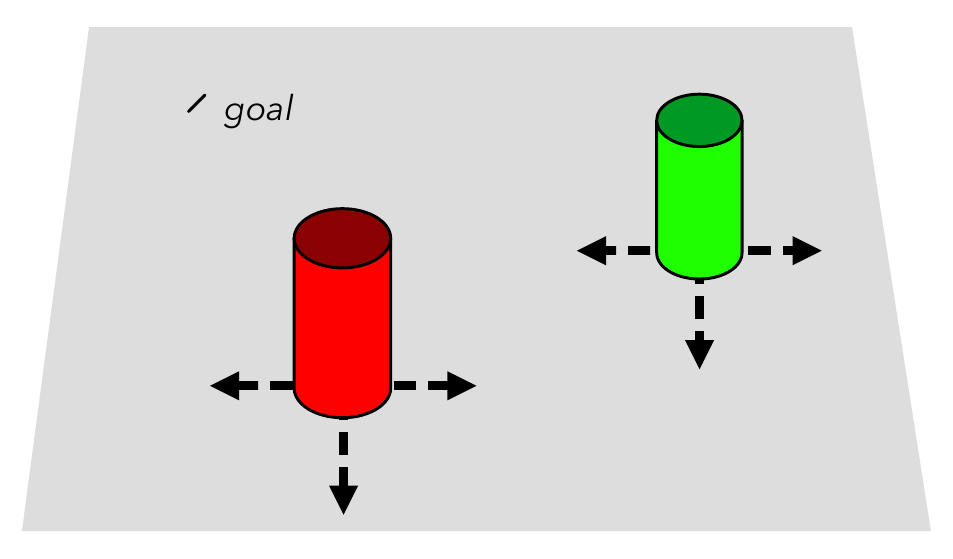}
        \caption{In this case, the agent is required to arrange the objects in the scene to satisfy a condition.}
        \label{fig:prep:goaltypes:b}
    \end{subfigure}
    \caption{Two goal environments with objects in 2-dimensional discrete space.}
    \label{fig:prep:goaltypes}
\end{figure}

There are two different types of goal satisfaction discussed in the relevant literature: the case where each state uniquely maps to a single goal~\citep{her} and the case where each state may map to multiple goals or none. To see a motivation for each case, consider the two environments shown in Figure~\ref{fig:prep:goaltypes}. In Figure~\ref{fig:prep:goaltypes:a}, the agent must move a single object to a goal position. It is then clear that, if the object is in some position $(x,y)$, it has satisfied the goal $(x,y)$. Hence, $\textsc{Sat}(s, g) \iff s = g$. In Figure~\ref{fig:prep:goaltypes:b}, the agent is asked to arrange the environment into a specific configuration to achieve the goal. There are multiple elements in the scene and a goal could be specified in terms of any object. For example, a state having the red cylinder at $(x_\mathrm{red}, y_\mathrm{red})$ and a green cylinder at $(x_\mathrm{green}, y_\mathrm{green})$ would simultaneously satisfy the goals ``move the red cylinder to $(x_\mathrm{red}, y_\mathrm{red})$'' \emph{and} ``move the green cylinder to $(x_\mathrm{green}, y_\mathrm{green})$. Approaches to goal-conditioned learning have focused on environments of either type~\citep{hir, her}.

\subsection{Reward Shaping}
\label{apdx:mdpreduction:goalrewards}

Designing a function $\mathcal{R}$ that incentivises an RL agent to perform desirable activities is known as \emph{reward shaping}~\citep{shaping}. It is non-trivial to select a suitable reward function in goal-conditioned learning, because the function must take account of the explicitly defined goal.

Whereas in traditional RL the agent aims to achieve the highest value of discounted reward, in goal-conditioned learning the aim becomes achieving the \emph{goal} with the \emph{assistance} of the reward function. To be precise, the set of actions required to maximise the total discounted reward must be exactly those actions that achieve the goal in the fewest steps. Further, since the goal is variable, the reward function must maintain this property regardless of the goal. We refer to this as the \emph{agreement} between a reward function $\mathcal{R}$ and a goal-satisfaction relation $\textsc{Sat}$.

There are two common approaches to determining reward in goal-conditioned environments.

\begin{itemize}
    \item \textbf{Binary Rewards}: The simplest reward for goal-conditioned environments is a binary reward. In this case, we define $\mathcal{R}$ as:
    \[
        \mathcal{R}((s, a, s^\prime), g) = \begin{cases*}
          0 & if $\textsc{Sat}(s^\prime, g)$ \\
          -1 & otherwise
        \end{cases*}
      \]
      
    It is easy to see how this approach agrees with any goal-conditioned environment. However, the signal is very sparse. Agents with sparse reward signals are known to learn more slowly~\citep{sparse}.
    
    \item \textbf{Distance-Based Rewards}: In some environments, a \emph{distance-like} metric $\mathcal{D}(s, g)$ exists between each state $s$ and goal $g$.
        
    The reward function can then be defined in two ways, for a \emph{monotonically increasing} function $f$:
    \begin{itemize}
        \item $\mathcal{R}((s, a, s^\prime), g) = f(-\mathcal{D}(s^\prime, g)) $ \hfill \emph{absolute credit}
        \item $\mathcal{R}((s, a, s^\prime), g) = f(\mathcal{D}(s, g) - \mathcal{D}(s^\prime, g)) $ \hfill \emph{relative credit}
    \end{itemize}
    
    If $\mathcal{G}=\mathcal{S}$ then $\mathcal{D}(s, g) = |s - g|$.

\end{itemize}

Each of these methods have been used in goal-conditioned environments to varying degrees of success. \citet{her} found that a binary reward signal outperformed each other option that was tried. For that reason, we utilise binary rewards.

    \begin{figure}
        \centering
        \includegraphics[width=6cm,keepaspectratio]{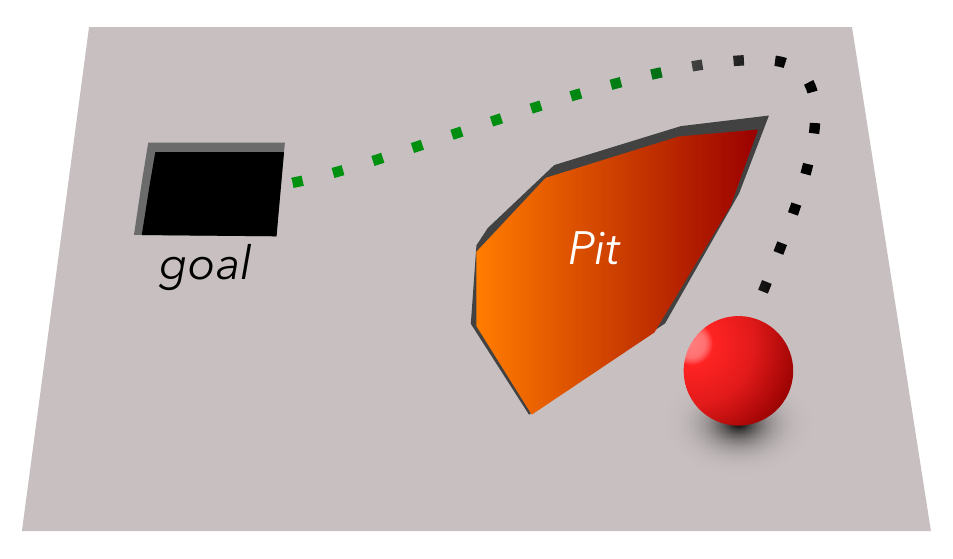}
        \caption{An example of a more complex goal-conditioned environment. In this case, the agent can fall into a pit, which will cause it to lose the game.}
        \label{fig:prep:goalenvharder}
    \end{figure}

\section{Choosing a `Mix-In' Probability} \label{apdx:mixin}

CHER and EHER select random goals to relabel from the trajectory (during training step $t$) with a certain probability $\mu(t)$, to reduce the risk of \emph{catastrophic forgetting}. If goal relabelling is considered as a form of curriculum, then $\mu(t)$ can be thought of as a way of satisfying the requirement to `mix in' easier tasks that the agent can already solve. \citet{execute} show that this problem is non-trivial for curricula, so it is impractical to aim for an optimal choice. We considered the following strategies:

\begin{itemize}
    \item Constant $\mu(t) = 0.5$. This maintains an equal inclusion of `easier' tasks with more useful ones.
    \item Increasing $\mu(t) \propto t$. In this case, as the agent gains proficiency, it samples more randomly. This \emph{could} help to protect against forgetting and increase the rate of learning by focusing primarily on the more interesting goals at the start and generalising to the others.
    \item Decreasing $\mu(t) \propto c - t$. A decreasing strategy could be effective by learning more broadly at the start, then focusing on goals that are harder to learn as training progresses.
    \item Constant $\mu(t) = 0.0$. The agent may be protected from forgetting through existing mechanisms such as the replay buffer and exploration strategy, so no random experiences are needed.
\end{itemize}

To evaluate these approaches, we ran 5 experiments using EHER with each strategy. For the increasing and decreasing cases, $\mu(t)$ varied linearly from $0.0$ to $1.0$ or from $1.0$ to $0.0$, respectively. The criterion for a successful model was that it maintained a high success rate. In order to appropriately determine the success rate, we recorded the median and quartile values across each of the 5 experiments as training progressed. Firstly, we investigated variable approaches to $\mu(t)$.

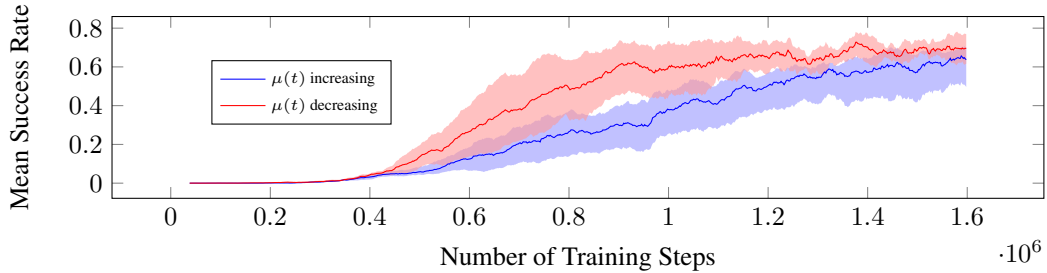
\begin{figure}
    \centering
    \begin{tikzpicture}
        \begin{axis}[
            xlabel={Number of Training Steps},
            ylabel={Mean Success Rate},
            width=\textwidth,
            height=4cm,
            legend style={at={(0.30,0.76)},anchor=north east},
            legend cell align={left},
            legend style={font=\tiny}]

            \addplot[blue] table [x=episode, y=median, col sep=comma] {data/processed_succ_down.csv};
            
            \addplot[red] table [x=episode, y=median, col sep=comma] {data/processed_succ_up.csv};
            
            \addplot [name path global=upper_a,draw=none] table[x=episode,y=uq,col sep=comma] {data/processed_succ_down.csv};
            \addplot [name path global=lower_a,draw=none] table[x=episode,y=lq,col sep=comma] {data/processed_succ_down.csv};
            \addplot [fill=blue!40,  fill opacity=0.6] fill between[of=lower_a and upper_a];

            \addplot [name path global=upper_b,draw=none] table[x=episode,y=uq,col sep=comma] {data/processed_succ_up.csv};
            \addplot [name path global=lower_b,draw=none] table[x=episode,y=lq,col sep=comma] {data/processed_succ_up.csv};
            \addplot [fill=red!40, fill opacity=0.6] fill between[of=lower_b and upper_b]; 
            
            \addlegendentry{$\mu(t)$ increasing}
            \addlegendentry{$\mu(t)$ decreasing}
            
        \end{axis}
    \end{tikzpicture}
    \caption{A comparison of the average success rates of agents using two different goal-relabelling strategies: increasing and decreasing. The average success rate for each agent is determined in $200$ randomly initialised test episodes and at the end of every $4000$ training steps. The decreasing strategy trains more quickly. In total, six agents were trained for each strategy. The solid line represents the median success rate and the shaded region represents the inter-quartile range.}
    \label{fig:result:upvsdown}
\end{figure}

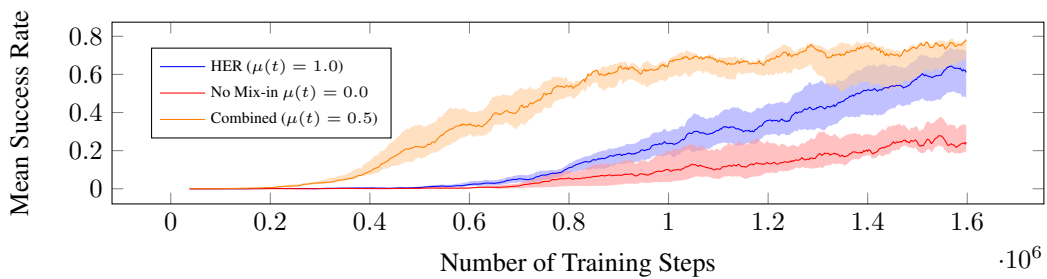
\begin{figure}
    \centering
    \begin{tikzpicture}
        \begin{axis}[
            xlabel={Number of Training Steps},
            ylabel={Mean Success Rate},
            width=\textwidth,
            height=4cm,
            legend style={at={(0.30,0.86)},anchor=north east},
            legend cell align={left},
            legend style={font=\tiny}]
            
            \addplot[blue] table [x=episode, y=median, col sep=comma] {data/processed_succ_low.csv};

            \addplot[red] table [x=episode, y=median, col sep=comma] {data/processed_succ_high.csv};
            
            \addplot[orange] table [x=episode, y=median, col sep=comma] {data/processed_succ_medium.csv};

            \addplot [name path global=upper_a,draw=none] table[x=episode,y=uq,col sep=comma] {data/processed_succ_low.csv};
            \addplot [name path global=lower_a,draw=none] table[x=episode,y=lq,col sep=comma] {data/processed_succ_low.csv};
            \addplot [fill=blue!40,  fill opacity=0.6] fill between[of=lower_a and upper_a];
            
            \addplot [name path global=upper_b,draw=none] table[x=episode,y=uq,col sep=comma] {data/processed_succ_high.csv};
            \addplot [name path global=lower_b,draw=none] table[x=episode,y=lq,col sep=comma] {data/processed_succ_high.csv};
            \addplot [fill=red!40, fill opacity=0.6] fill between[of=lower_b and upper_b];

            \addplot [name path global=upper_c,draw=none] table[x=episode,y=uq,col sep=comma] {data/processed_succ_medium.csv};
            \addplot [name path global=lower_c,draw=none] table[x=episode,y=lq,col sep=comma] {data/processed_succ_medium.csv};
            \addplot [fill=orange!40, fill opacity=0.6] fill between[of=lower_c and upper_c];
            
            \addlegendentry{HER ($\mu(t)=1.0$)}
            \addlegendentry{No Mix-in $\mu(t)=0.0$}
            \addlegendentry{Combined ($\mu(t)=0.5$)}
            
        \end{axis}
    \end{tikzpicture}
    \caption{A comparison of the success rate of agents during training. The most successful approach is the case where $\mu=0.5$---picking goals randomly with a probability of $0.5$. The blue line represents traditional HER. In total, six agents were trained for each strategy. The solid line represents the median success rate and the shaded region represents the inter-quartile range.}
    \label{fig:result:1vs0}
\end{figure}

The success rate during training for the cases where $\mu(t)$ increases and decreases is shown in Figure~\ref{fig:result:upvsdown}. The agent trains more quickly in the case where $\mu(t)$ is \emph{decreasing}. Figure~\ref{fig:result:1vs0} compares HER, $\mu(t)=1.0$ and $\mu(t)=0.5$. Note that $\mu = 1.0$ represents HER. Clearly, the most effective choice is $\mu(t)=0.5$. While it was anticipated that $\mu(t)=0.0$ might result in the agent forgetting some experiences, it was surprising to see that it actually performed the worst among all of the strategies that were tested.

A particularly useful insight into the failure of $\mu(t)=0.0$ can be found by examining the values for TD-error during training. These are collected by taking the mean TD-error over each goal achieved in the trajectory of each episode during training. When aggregated, the mean TD-error is approximately indicative of how well an agent has learned the environment at any point. The TD-error was recorded during training for each of the 5 models for every strategy.

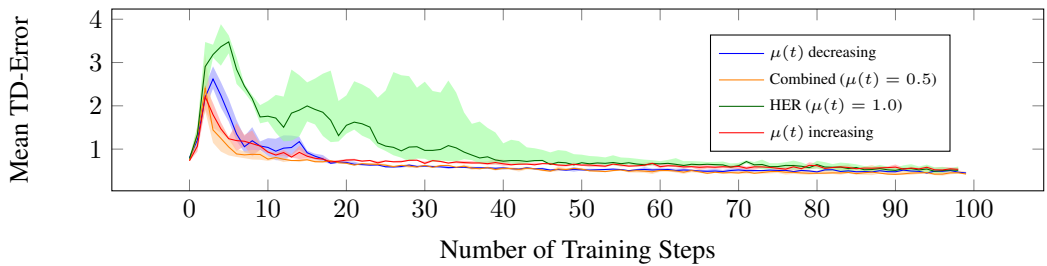
\begin{figure}
    \centering
    \begin{tikzpicture}
        \begin{axis}[
            xlabel={Number of Training Steps},
            ylabel={Mean TD-Error},
            width=\textwidth,
            height=4cm,
            legend style={at={(0.9,0.86)},anchor=north east},
            legend cell align={left},
            legend style={font=\tiny}]

            \addplot[blue] table [x=s, y=median, col sep=comma] {data/error_processed_succ_up.csv};

            \addplot[orange] table [x=s, y=median, col sep=comma] {data/error_processed_succ_medium.csv};

            \addplot[black!60!green] table [x=s, y=median, col sep=comma] {data/error_processed_succ_low.csv};

            \addplot[red] table [x=s, y=median, col sep=comma] {data/error_processed_succ_down.csv};

            \addplot [name path global=upper_a,draw=none] table[x=s,y=uq,col sep=comma] {data/error_processed_succ_up.csv};
            \addplot [name path global=lower_a,draw=none] table[x=s,y=lq,col sep=comma] {data/error_processed_succ_up.csv};
            \addplot [fill=blue!40,  fill opacity=0.6] fill between[of=lower_a and upper_a];

            \addplot [name path global=upper_b,draw=none] table[x=s,y=uq,col sep=comma] {data/error_processed_succ_down.csv};
            \addplot [name path global=lower_b,draw=none] table[x=s,y=lq,col sep=comma] {data/error_processed_succ_down.csv};
            \addplot [fill=red!40, fill opacity=0.6] fill between[of=lower_b and upper_b];

            \addplot [name path global=upper_c,draw=none] table[x=s,y=uq,col sep=comma] {data/error_processed_succ_medium.csv};
            \addplot [name path global=lower_c,draw=none] table[x=s,y=lq,col sep=comma] {data/error_processed_succ_medium.csv};
            \addplot [fill=orange!40, fill opacity=0.6] fill between[of=lower_c and upper_c];

            \addplot [name path global=upper_d,draw=none] table[x=s,y=uq,col sep=comma] {data/error_processed_succ_low.csv};
            \addplot [name path global=lower_d,draw=none] table[x=s,y=lq,col sep=comma] {data/error_processed_succ_low.csv};
            \addplot [fill=green!40, fill opacity=0.6] fill between[of=lower_d and upper_d];

            \addlegendentry{$\mu(t)$ decreasing}
        
            \addlegendentry{Combined ($\mu(t)=0.5$)}
            \addlegendentry{HER ($\mu(t)=1.0$)}
            
            \addlegendentry{$\mu(t)$ increasing}

        \end{axis}
    \end{tikzpicture}
    \caption{The TD-error of episodes as training progresses for various mix-in strategies. The mean TD-error of all relabelling options is recorded after each training episode. A rolling average with a square window of length $20$ is used to smooth the data. Six agents are trained for each policy. The solid line represents the median TD-error among the agents and the shaded region represents inter-quartile range. Of the strategies displayed, the agents trained using HER experience the highest TD-error at the start of training. A higher TD-error suggests lower familiarity with the environment.}
    \label{fig:result:39tderror}
\end{figure}

Plots representing TD-error over time are illustrated for different strategies in Figure~\ref{fig:result:39tderror}. From this, we can see that any of the approaches where $\mu(t)$ increases, decreases, or is fixed at $0.5$ lead to a quicker reduction in uncertainty than traditional HER. Specifically, this biggest improvement over HER occurs at the start of training, which corresponds with the success rate plot in Figure~\ref{fig:result:1vs0}.

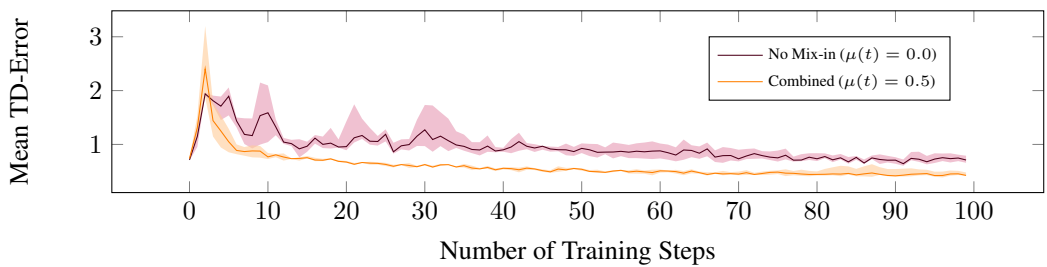
\begin{figure}
    \centering
    \begin{tikzpicture}
        \begin{axis}[
            xlabel={Number of Training Steps},
            ylabel={Mean TD-Error},
            width=\textwidth,
            height=4cm,
            legend style={at={(0.9,0.86)},anchor=north east},
            legend cell align={left},
            legend style={font=\tiny}]

            \addplot[black!60!purple] table [x=s, y=median, col sep=comma] {data/error_processed_succ_high.csv};

            \addplot[orange] table [x=s, y=median, col sep=comma] {data/error_processed_succ_medium.csv};
            
            \addplot [name path global=upper_a,draw=none] table[x=s,y=uq,col sep=comma] {data/error_processed_succ_high.csv};
            \addplot [name path global=lower_a,draw=none] table[x=s,y=lq,col sep=comma] {data/error_processed_succ_high.csv};
            \addplot [fill=purple!40,  fill opacity=0.6] fill between[of=lower_a and upper_a];

            \addplot [name path global=upper_c,draw=none] table[x=s,y=uq,col sep=comma] {data/error_processed_succ_medium.csv};
            \addplot [name path global=lower_c,draw=none] table[x=s,y=lq,col sep=comma] {data/error_processed_succ_medium.csv};
            \addplot [fill=orange!40, fill opacity=0.6] fill between[of=lower_c and upper_c];

            \addlegendentry{No Mix-in ($\mu(t)=0.0$)}
            \addlegendentry{Combined ($\mu(t)=0.5$)}
            
        \end{axis}
    \end{tikzpicture}
    \caption{The TD-error of agents trained using the `combined' mix-in policy $\mu=0.5$ from Figure~\ref{fig:result:39tderror} and using the `no mix-in' policy $\mu=0.0$. The solid lines represent median values taken across agents and the shaded regions represent the inter-quartile range. Despite only training on those experiences with a high TD-error, the `no mix-in' policy actually records a higher TD-error among samples from the environment than the `combined' approach.}
    \label{fig:result:39tderrorother}
\end{figure}

Despite training focusing \emph{exclusively} on the examples with the highest TD-error, the case where $\mu(t)=0.0$ does not lead to the lowest TD-error in the long term. Figure~\ref{fig:result:39tderrorother} compares the TD-error in the cases of $\mu(t)=0.0$ and $\mu(t)=0.5$. By once again considering goal relabelling as a curriculum, the failure of $\mu(t)=0.0$ corresponds to the observation made by~\citet{execute} that easier tasks must be `mixed in'. In this case, a lower TD-error value being associated with a goal suggests that an agent would find it `easier' to achieve. Hence, by randomly selecting from these `easier' goals, EHER is able to `mix in' easier `tasks' and make learning more effective.

\section{DigitFlip Model Success Rate} \label{apdx:model_succ_rate}

\begin{figure}[h]
    \centering
    \includegraphics[width=.4\textwidth, keepaspectratio]{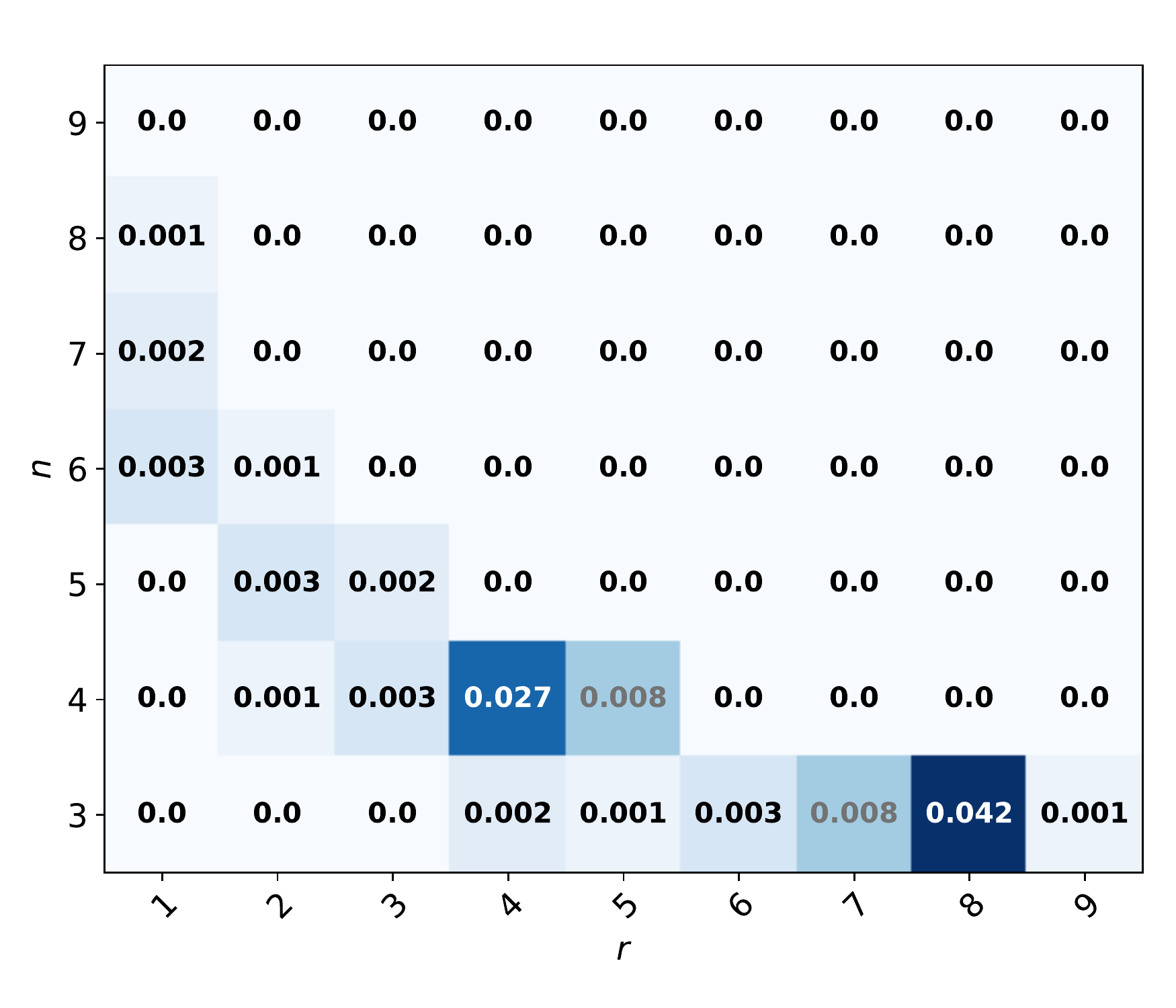}
    \caption{The variance between the success rates of the $5$ models trained for each value of $n$ and $r$ in Figure~\ref{fig:dig_all}. Lower values indicate that the models had more similar success rates. Since almost all of the values are very close to $0$, the mean success rate is a reliable metric to use when determining how difficult an environment is. Darker colours represent higher numerical values.}
    \label{fig:dig_varia}
\end{figure}

\section{Experimental Details} \label{apdx:exp}

Both \texttt{DigitFlip} and \texttt{CLEVR-Play} were implemented as \emph{OpenAI} Gym environments. The implementations of EHER, CHER and IGOAL have also been integrated with the \texttt{stable-baselines} library\footnote{\url{https://stable-baselines.readthedocs.io/en/master/}}. Models are trained using the default \texttt{stable-baselines} (Double) DQN implementation, with an epsilon-greedy exploration strategy. The value of epsilon is initialised to $1.0$ and decays exponentially by a factor of $0.993$ on each epoch to $0.1$. The Q-value update step uses a discount factor of $0.9$.

For each training epoch of \texttt{CLEVR-Play}, there are 50 \emph{cycles}. At the start of each cycle, the parameters of the target network are copied into the primary network. When agents are trained on \texttt{DigitFlip}, the parameters are copied every 4000 steps. Updates to the primary network are applied using \emph{Polyak averaging} with $\tau=0.05$.

The learning rate used in all experiments is $5 \times 10^{-4}$. The buffer size used for experience replay is $2 \times 10^6$. The batch size is 64.

\end{document}